\documentclass{svproc}
\usepackage[hyphens]{url}
\usepackage{mathrsfs}  
\usepackage{amssymb,amsmath}
\usepackage{graphicx}
\usepackage{algorithm}  
\usepackage{algorithmic}
\usepackage{subfigure}
\usepackage{color}
\usepackage{graphicx}
\usepackage{url}
\usepackage{wrapfig}
\usepackage{picinpar}
\usepackage{cutwin}
\usepackage{hyperref}
\usepackage{amsfonts,amssymb}

\begin{document}
\mainmatter              
\title{A Regressive Convolution Neural network and Support Vector Regression Model for Electricity Consumption Forecasting}
\titlerunning{RCNN-SVR}  
%
\author{Youshan Zhang\inst{1} \and Qi Li\inst{2}}
\authorrunning{Youshan Zhang et al.} 
%
\tocauthor{Youshan Zhang, Qi Li}
\institute{Computer Science and Engineering, Lehigh University, Bethlehem, PA 18015, USA\\
\email{yoz217@lehigh.edu},
\and
Department of Automation, BOHAI University, Liaoning, Jinzhou, 121013, China \\
\email{liqi199507@gmail.com}}

\maketitle              

\begin{abstract}
Electricity consumption forecasting has important implications for the mineral companies on guiding quarterly work, normal power system operation, and the management. However, electricity consumption prediction for the mineral company is difficult since electricity consumption can be affected by various factors. The problem is non-trivial due to three major challenges for traditional methods: insufficient training data, high computational cost and low prediction accuracy. To tackle these challenges, we firstly propose a Regressive Convolution Neural Network (RCNN) model, but RCNN still suffers from high computation overhead. Then we utilize RCNN to extract features from data and Regressive Support Vector Machine (SVR) trained with features to predict the electricity consumption. The experimental results show that RCNN-SVR model achieves higher accuracy than using the traditional RCNN or SVM alone. The MSE, MAPE, and CV-RMSE of RCNN-SVR model are 0.8564, 1.975\%, and 0.0687\% respectively, which illustrates the low predicting error rate of the proposed model.
\end{abstract}
\section{Introduction}

The electricity consumption of large enterprises has been a major factor of the cost control and the operational efficiency. Specifically, mineral companies consume large quantities of electricity in the coal production process daily. The electricity consumption forecasting has important implications for the mineral companies on guiding quarterly work, the normal power system operation and power management. Besides, the prediction accuracy of electricity consumption directly determines the power construction, network planning and the planning of electricity marketing strategies \cite{Abdollah,Song,Fazil,Zhang}. Therefore, predicting the electricity consumption accurately is demanded and crucial to mineral companies. 

Since the complicated dynamic of the electrical power system, it is difficult to establish an explicit model. Many traditional methods are applied to predict the electricity consumption, such as Gray prediction, regression analysis, time series, artificial neural network (ANN), support vector machine (SVM) \cite{Fazil,Diyar,Vincenzo,RE,L,Shuai}. However, these methods have their respective disadvantages. For example, traditional ANN train data are mostly based on the gradient, and it may fail into local minimum easily \cite{Zhang}. One common limitation is that these methods are strongly depended on the number of training data, which discover the relationship between predictive value and model. Also, some statistical analysis models such as Kalman filters, and Autoregressive Integrated Moving Average (ARIMA) \cite{Chaoqing,Ted,HM} were also applied in electricity consumption prediction. However, they still have constraints of insufficient data size. In \cite{Yi}, Hu presented a neural-network-based gray prediction (NNGM(1,1)) method, which can overcome the limitation of the traditional gray prediction method. It can easily determine the developing coefficient and control variables in the gray prediction model. Therefore, NNGM(1,1) can improve load forecasting accuracy. Similarly, in \cite{Song}, Song et al. modified the gray prediction method and proposed a rolling gray prediction(NOGM(1,1)) model. \cite{Song} overcame the deficiencies of fixed structure and poor adaptability in the original gray prediction model. The empirical results showed the NOGM(1,1) model has higher prediction accuracy than original gray prediction model. However, the prediction accuracies of these methods are still not satisfying. 

The major challenge is that electricity consumption prediction of the mineral company is different from the traditional electricity load prediction since mineral company electricity consumption is affected by various factors (e.g., ore grade, processing quantity of the crude ore, Ball milling fill rate). Conventional methods only consider the electricity values and ignore the influential factors. Therefore, it is necessary to build a new model that not only considers electricity values and influential factors but predicts the monthly electricity consumption of mineral company. In this paper, we will solve three issues by our proposed electricity consumption prediction model: (1) reduce the computational cost; (2) train the model with limited data; and (3) improve the prediction accuracy. Convolution Neural Network (CNN) \cite{Alex,Jonathan,Nal} has become a popular method for solving image classification, segmentation, and regression problem recently. However, there is no such a Regressive CNN (namely RCNN, ending with a regression layer) architecture for predicting electricity consumption of mineral company.

In this study, we present a new electricity consumption forecasting model based on regressive convolution neural network and support vector regression (RCNN-SVR). Compared with traditional methods, the RCNN model is capable of extracting more representative features of history electricity consumption data, while SVR model can reduce the computation overhead. The forecasting accuracy of the proposed model is higher than several baseline models such as BP neural network and SVM \cite{Fazil}. There are two major contributions of this paper: (1) build the RCNN-SVR architecture to predict the electricity consumption of electricity; (2) compare prediction performances of our model with several baseline forecasting methods. We describe the RCNN and the SVR model, and introduce the model architecture in section \ref{sec:method}. Experiments are conducted to verify our model and the comparisons with previous methods are available in section \ref{sec:results}. Based on results in section \ref{sec:results}, we discuss the experimental results, make a conclusion and explore future work in section \ref{sec:conclusion}.

\section{Methodologies} \label{sec:method}
In this section, we first introduce the regressive convolution neural network (RCNN), and support vector regression(SVR) model, separately. Then, we present our RCNN-SVR architecture for predicting electricity consumption.
\subsection{Data prepossessing}\label{sec:data}
The electricity consumption data was collected from a mineral company in Liaoning province, China. It contained the monthly electricity consumption from 2012 to 2017 (only two months data are provided in 2017) with total 62 months. We split the data into training data and testing data. Testing data are not used during the training process. Training data contain $8 \times 50$ influential factors(IFs) 8 is eight IFs of each month, and 50 is the number of month. $50 \times 1$ true electricity consumption values(EVs). Testing data contain $8 \times 12$ IFs, $12 \times 1$ true EVs. For the input for RCNN, and RCNN-SVR model, we reshape the influential factors into a 4-D array, for example, influential factors change into $8\times 1 \times 1 \times 50$ for training and testing dataset, $8, 1$ and $1$ represents for length, height, and depth.
\subsection{RCNN Architecture} \label{sec:rcnn}

 We first propose a regressive convolution neural network model(RCNN, shows in Fig. 1), which is similar to DeepEnergy in \cite{Ping}. But our RCNN model has fewer layers because of limited data, the input is influential factors (IFs), and the last layer is regression layer which represents the electricity consumption values(EVs). In this network, it contains two main steps: feature extraction, and prediction. It only has eight layers. The feature extraction is performed by two convolution layers (Conv1, Conv2), and two max-pooling layer, (Maxpool1, Maxplool2), one rectified linear units (ReLU) layer, and one normalization (Norm) layer. The prediction step consists of a fully-connected layer and a regression layer. The input layer is comprised of $8\times 1$ influential factors (one month), Conv1 and Conv2 have the filter size ($F$) of $1\times 1$, and filter number ($N$) 25 with padding size ($P$) 0; Maxpool1 and Maxpool2 have the stride size ($S$) of $2\times 2$. Therefore, after the max-pooling layer, the dimension of feature map is divided by 2. The ReLU layer reduces the number of epochs to achieve the training error rate greater than traditional tanh units. The normalization layer increases generalization and reduces the error rate. Also, ReLU and normalization layer does not change the size of the feature map. The pooling layers summarize the outputs of adjacent pooling units\footnote{Source code is available at: \url{https://github.com/heaventian93/A-Regressive-Convolution-Neural-Network-and-Support-Vector-Regression-Model-for-Electricity-Consumpt.}}. 
 
 One of the most obvious merits of RCNN is more features can be extracted from different layers. With more features, we can easily build the relationship between model and the predicted value. For example, if the input size ($I$) is $8\times 1$, we assume that feature map size is $8\times 1$. In Conv layer, the feature map size can be calculated as: $((I-F+2P)/S+1)\times N$. And feature map size is equal to $I/S\times N$ in max-pooling layers. In the Conv1 layer, the feature map size is: $((8-1)/1+1) \times 25=8\times 25$; the feature map size in Maxpool1 is $(8/2)\times 25=4\times 25$. Again, the feature size is: $((4-1)/1+1)\times 25=4\times 25$ in the Maxpool2 layer, and feature size becomes: $(4/2)\times 25=2\times 25$ in the Maxpool2 layer. The total number of features is increased (50 in maxpooling2 layer v.s. 8 in input layer), and this is one reason why RCNN can generate a better-predicted result than other neural networks which only use input data as feature map.  

\begin{figure}[H]
\centering
\includegraphics[scale=0.35]{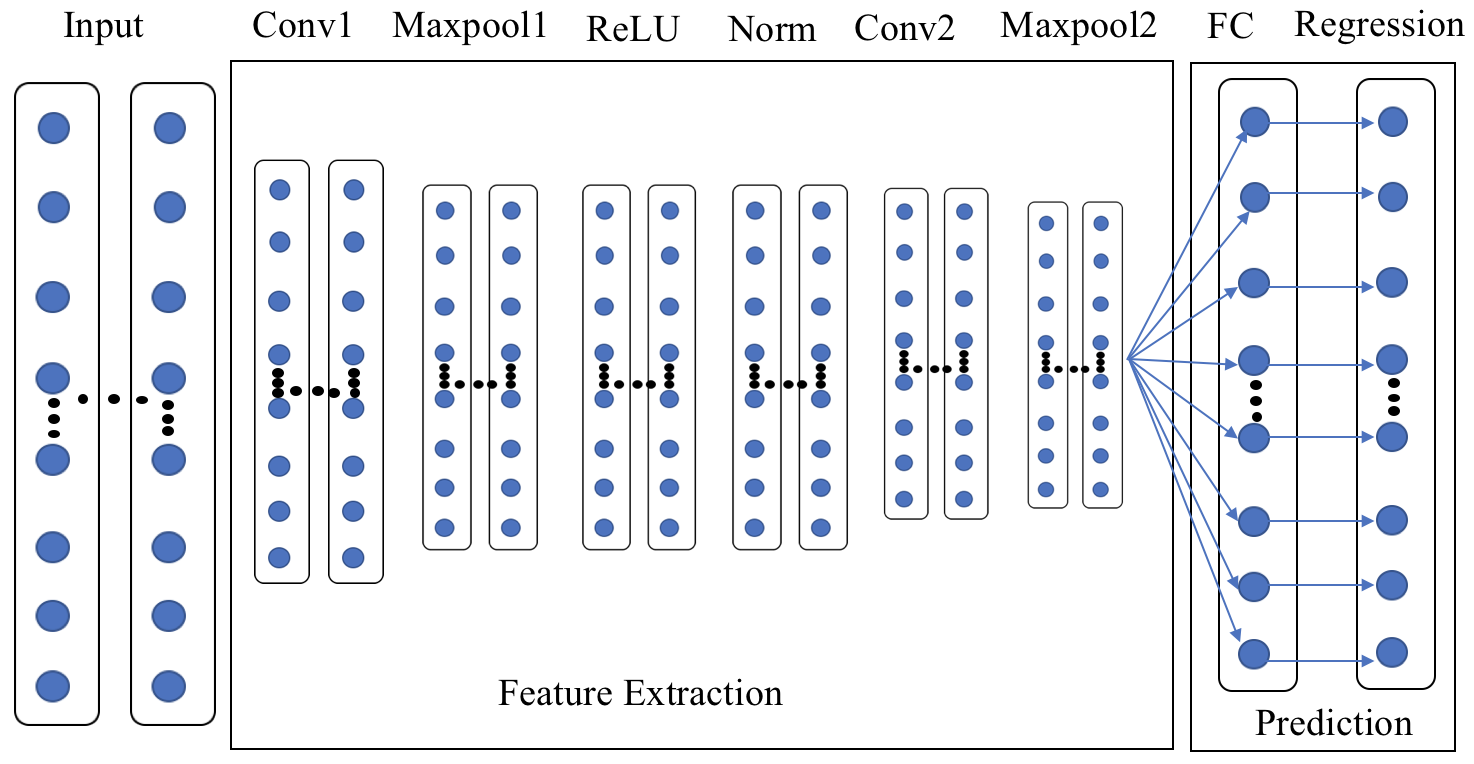}\\
\caption{The RCNN structure, it contains only eight layers (to prevent the overfitting of limited data), the input layer is the: influential factors. And the regression layer generates electricity consumption values. In the training stage, the RCNN will extract features of the influential factors, and check if the MSE is convergent.  By using the trained RCNN classifier, we can predict the electricity consumption values of test data. (Conv: convolution layer, NA: normalization layer, FC: fully-connection layer.) } 
\end{figure}

 \subsubsection{Electricity prediction using RCNN}
   As shown in Fig. 1, with more features extracted in the Maxpooll2 layer, we will connect it into FC layer and flat all features into one dimension. In the training stage, the input size is: $8\times 50$. The size of the fully-connected layer is $50 \times 1$, and it has the same size as the regression layer, and this why points in FC layer are only connected to one point in regression layer. During the training process, if the desired Mean Square Error (MSE) is not reached in the current epoch, the training will continue until the maximal number of epochs or desired MSE is reached. On the contrary, if the maximal number of epochs is reached, then the training process will stop regardless the MSE value. Final performances are evaluated to demonstrate feasibility and practicability of the proposed method. During the test stage, we input the test data set $8\times 12$, and by using the training RCNN model, we can predict the electricity consumption of each month. 
\subsection{SVR}
The original linear support vector machine (SVM) is proposed for binary classification problem. Given data and its labels: $(x_{n}, y_{n})$, $n=1,..., N$, and $y_{n}\in \{-1, +1\}$. It aims to  optimize following  equation: 
\begin{equation}
\begin{aligned}
          \min_{\vec w,b}  \frac 1 2 \| \vec w \|^2 + \lambda \sum_n \xi_n^2 
    \quad\text{s.t.}  \\\quad  y_n (\vec w^T \vec x_n + b) \geq 1 - \xi_n \quad (\forall n), \quad 
                 \xi_n \geq 0 \quad(\forall n),
\end{aligned}
\end{equation}
where $\lambda$ controls the width of margin (smaller margin with smaller $\lambda$);  $\xi_{n}$ is a non-negative slack variable and penalizes data points which against the margin; $b$ is the bias. 

Linear SVM  can also be used as a  regression method (called SVR), there are few minor differences comparing with SVM for classification problem. First of all, the output of SVR is a continuous number, but not the classes in the classification problem. Besides, there is a margin of tolerance $\varepsilon$ in the SVR. However, the main idea is always the same: minimize the error and maximize the margin. Fig. 2 describes the one-dimensional SVR, it aims to optimize following constrained function:
\begin{equation}
\begin{aligned}
 \min_{\vec w,b}  \frac 1 2 \| \vec w \|^2 + \lambda \sum_n (\xi_n + \xi_{n}^{*})
     \quad \text{s.t.} \quad  y_n -(\vec w^T \vec x_n + b) \leq \varepsilon + \xi_n \\ \quad  y_n -(\vec w^T \vec x_n + b) \leq \varepsilon + \xi_{n}^{*} \quad (\forall n), \quad
\xi_n,\  \xi_{n}^{*} \geq 0 \quad(\forall n),
\end{aligned}
\end{equation}
where  $\xi_{n}^{*}$ is another non-negative slack variable \cite{Smola,Debasish,Yichuan}. 

\begin{figure}[H]
\centering
\includegraphics[scale=0.4]{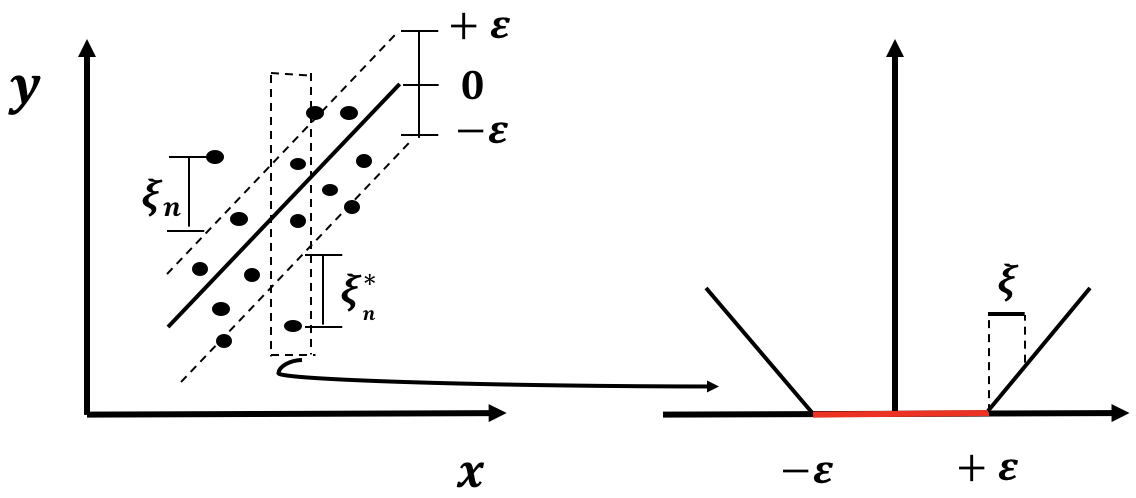}\\
\caption{The one-dimensional linear SVR with $\varepsilon$ intensive band. Only the points which is out of $+\varepsilon$ and $-\varepsilon$ bound contribute to the cost. The dashed rectangle can be alternative visualized in the left image. Figure is modified from \cite{Smola}. } 
\end{figure}

\subsubsection{Electricity prediction using SVR}
To apply SVR method in predicting the electricity consumption of mineral company, we use SVR classifier to train the eight factors and predict the electricity consumption using the trained classifier. The SVR structure is shown in Fig. 3. In training stage, we train the SVR classifier using IFs, and we compare the predicting electricity value with true EVs and check whether the model is convergent; if not, the training stage will execute again. During test stage, we use IFs from test data set and predict electricity value. 

\subsection{RCNN-SVR}

Inspired by the RCNN and SVR, we combine the deep neural network with SVR and design an RCNN-SVR model. Specifically, we train SVR classifier using features, which extracted from RCNN, then predict the electricity consumption using trained SVR classifier. Different from above RCNN architecture, we add more layers in the RCNN part to get more useful features. The RCNN-SVR architecture is shown in Fig. 4. Different from single RCNN and SVR model, RCNN-SVR combines the advantages of these two methods. RCNN-SVR can extract more features and use the less computational time to train the model. In our RCNN-SVR model, it also contains two steps: the feature extraction step is from RCNN model, and predicting step is from SVR model. Also, to extract the features, we fine-tuned the network. Different from the number of layers in RCNN, we add another Conv3 and Maxpool3 layer. To reduce error and prevent the overfilling, we use the drop out strategies, which adds a droppoutlayer after the  Maxpool3 layer. For three Conv layers, the fitter size is $1\times1$, and the filter number are: $20,  25$, and $50$, respectively. For three Maxpool layers, the stride size is $2\times2$.  Besides, we removed the last two layers (FC and regression layer), since we could not extract significant features from these two layers. The feature size of last dropout layer is the same as the feature map size in the Maxpool2 layer of RCNN. But the feature map is different; there is more information in feature map of RCNN-SVR model. As shown in Fig. 5 and Fig. 6, the feature map of RCNN-SVR model (both training and testing data) has more features than 8 layers RCNN model in section \ref{sec:rcnn}. With more features extracted in RCNN model, it will provide enough information for SVR model to train the features. Further, we can build a better relationship between features and actually electricity consumption values.

\subsubsection{Electricity prediction using RCNN-SVR}
 To apply the RCNN-SVR model in predicting the electricity consumption of mineral company, we use RCNN to extract features of eight IFs and predict the electricity consumption using the trained SVR classifier. The RCNN-SVR structure is shown in Fig. 4. As shon in Alg.\ref{alg:Alg.1}, in training stage, we train the SVR classifier using features from RCNN model, and we compare the predict electricity value with true EVs and check whether the model is convergent. If not, the training stage will execute again. During test stage, we use IFs from test data set and predict electricity consumption values.

 \begin{figure}[H]
\centering
\includegraphics[scale=0.37]{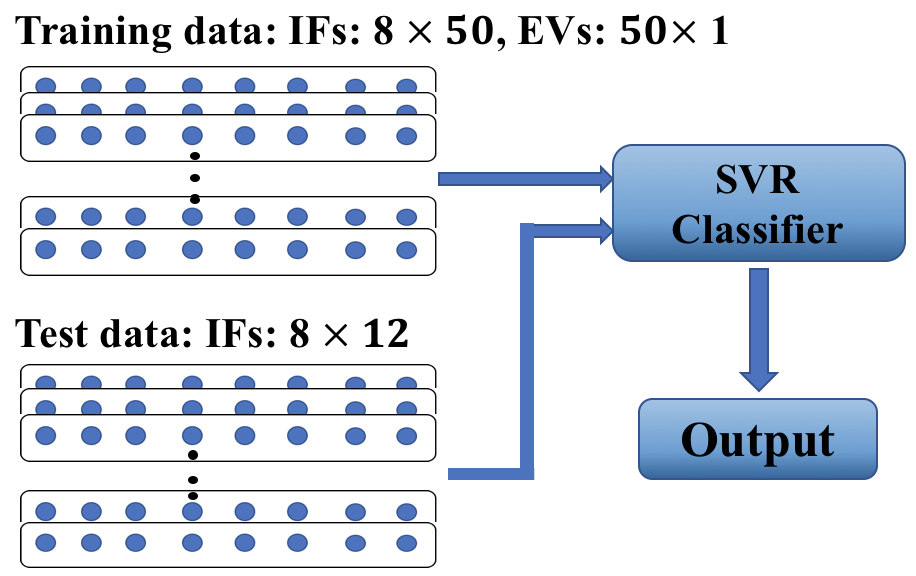}\\
\caption{The SVR structure, IFs: influential factors (eight factors with fifty months), EVs: electricity values. In the training stage, the SVR classifier trains the IFs, and check if the model is convergent, if not, then SVR model will train again. We can predict the electricity values using the trained SVR classifier.  } 
\end{figure}

\begin{figure}[H]
\centering
\includegraphics[scale=0.33]{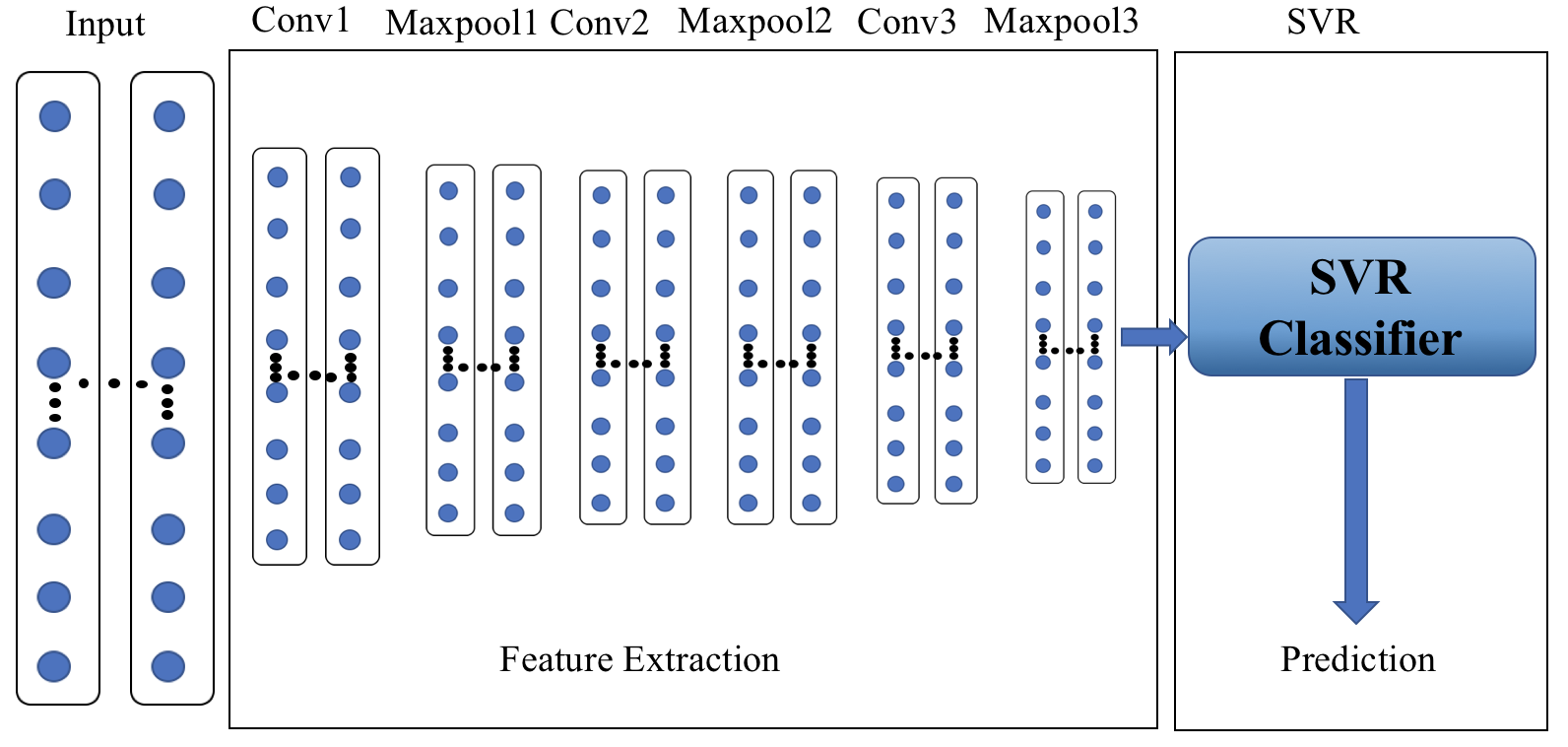}\\
\caption{The RCNN-SVR structure, which combines partial model from RCNN and SVR. Different from RCNN and SVR, RCNN-SVR extract features of data with more layers, and it trains SVR with more extracted features. By using RCNN-SVR model, we can extract more useful information and use less time to train the SVR classifier, and predict the electricity consumption values.  }
\end{figure}

\begin{algorithm}[H]
\caption{Electricity prediction using RCNN-SVR model}\label{alg:Alg.1} 
\textbf{Training Stage:}\\
Input data: influential factors: 8$\times$50, and electricity values: 1$\times$50\\
Extract the features from RCNN, and train SVR classifier\\

\textbf{Testing Stage:}\\
Input data: influential factors: 8$\times$12 \\
Predict electricity values of testing data using SVR classifier from testing stage\\
Calculate the accuracy of predicted results

\end{algorithm}

\begin{figure}
\begin{minipage}[t]{0.5\linewidth}
\centering\includegraphics[scale=0.38]{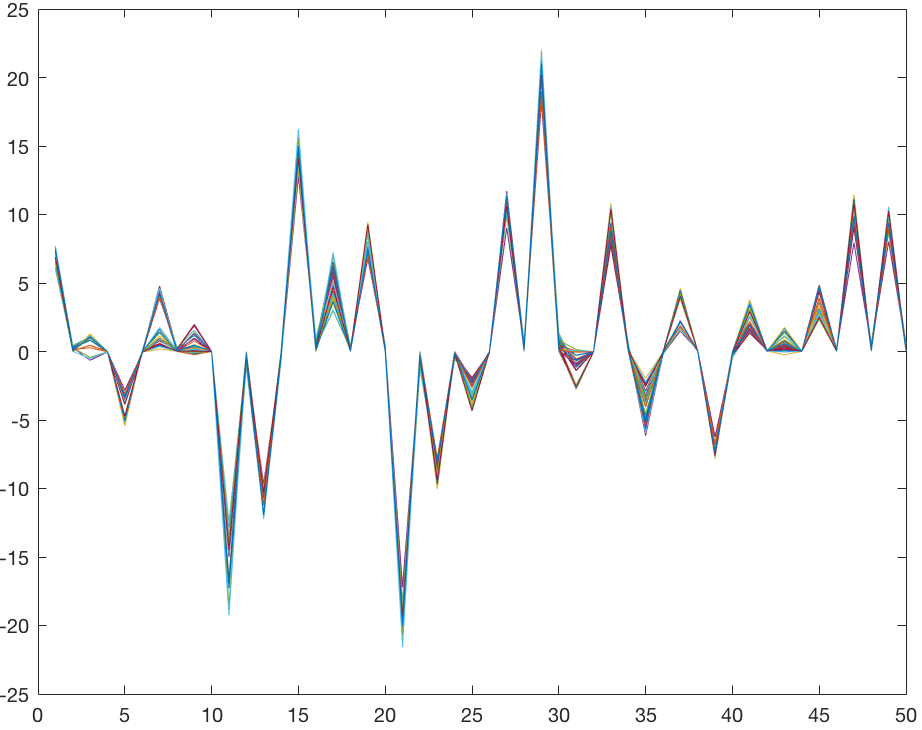}
\end{minipage}
\begin{minipage}[t]{0.5\linewidth}
\centering
\includegraphics[scale=0.38]{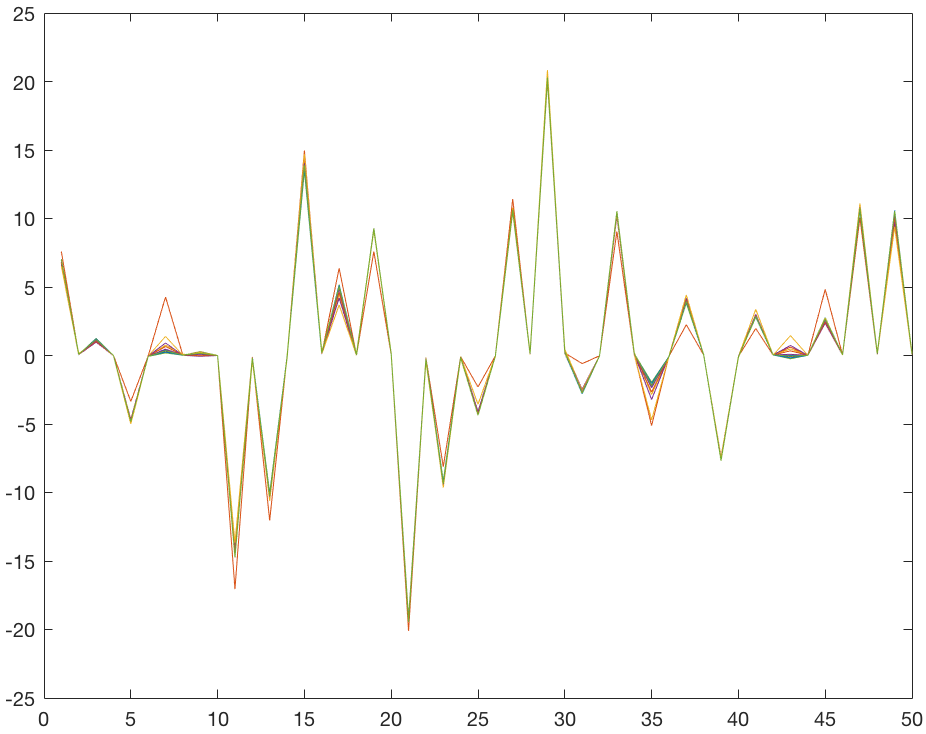}
\end{minipage}
\caption{The feature maps of training data (left one) and testing data (right one) from RCNN model. The x-axis is the number of features and y-axis is the range of features.  These features are extracted from the maxpooling2 layer in RCNN model.  } \label{fig:cont}
\end{figure}

\begin{figure}[H]
\begin{minipage}[t]{0.5\linewidth}
\centering
\includegraphics[scale=0.38]{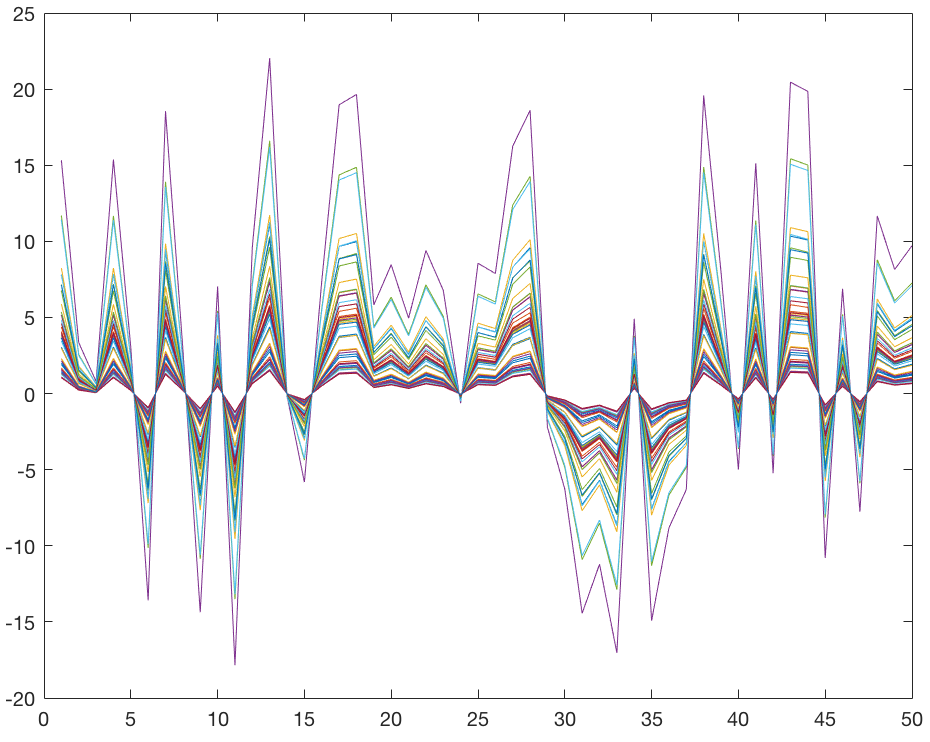}
\end{minipage}
\begin{minipage}[t]{0.5\linewidth}
\centering
\includegraphics[scale=0.38]{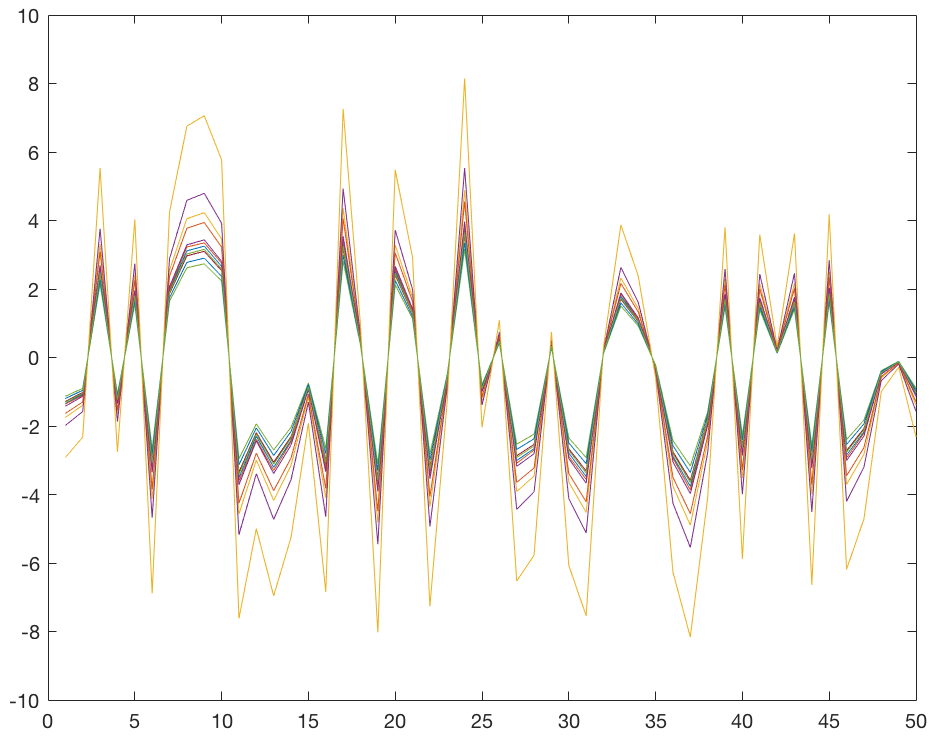}
\end{minipage}
\caption{The feature maps of training data (top one) and testing data (bottom one) from RCNN-SVM model. The x-axis is the number of features and y-axis is the range of features. We train the features of training data, and predict the electricity consumption of test data by using the extracted test features. Features are extracted from the maxpooling3 layer in RCNN-SVR model. There are 50 lines in the top image which are corresponding to the number of training data, and there are 12 lines in bottom image, which are corresponding to the number of testing data. Each line has different shapes which illustrate the difference of data.  We could find that RCNN-SVM model have more features than RCNN model. And this can be a reason that predicting results of RCNN-SVR model are better than RCNN model.}
\end{figure}

\section{Results} \label{sec:results}

In the experiment, we use data which are provided by a mineral company.  Besides, the training data are the electricity consumption values of past 50 months, and the test data are 12 months electricity consumption values. The data were processed in section \ref{sec:data}. Fig. 7 and Fig. 8 is the comparison predicting result of RCNN-SVR model with RCNN, SVR, MPSO-BP, and DeepEnergy. In Fig. 7, the vertical axes represent the electricity consumption (kWh), and the
horizontal axes denote different test months. According to the results in Fig.8, RCNN-SVR model has the highest accuracy among all models.

\subsection{Evaluation of model accuracy}

To evaluate the performance of predicting results, we employ three evaluation functions: Mean Standard Error (MSE), Mean Absolution Percentage Error (MAPE) and Cumulative Variation of Root Mean Square Error (CV-RMSE) \cite{Ping}. And these evaluation functions are defined in equation (3), where $y_{i}$ is the true electricity value, $\hat{y_{i}}$ is the predicting value, $N$ represents the data size.

\begin{equation}\label{eq:gtau}
\begin{aligned}
    \text{MSE}= \frac{\underset{i=1}{\overset{N}{\sum}}  (y_{i}-\hat{y}_{i})}{N},  \text{MAPE}=\frac{1}{N}\sum_{i=1}^{N} \left| \frac{(y_{i}-\hat{y}_{i})}{y_{i}} \right|, \\ \text{CV-RMSE}=\frac{\sqrt{\frac{1}{N}\underset{i=1}{\overset{N}{\sum}}  ( \frac{(y_{i}-\hat{y}_{i})}{y_{i}})^{2}}}{\frac{1}{N}\underset{i=1}{\overset{N}{\sum}} y_{i}}
\end{aligned}
\end{equation}

The comparison results of four methods are shown in table \ref{tab:my_label}. As shown in table \ref{tab:my_label}, the MAPE and CV-RMSE of the RCNN-SVR model are the smallest, and the goodness of error is the best among all models, namely, MSE, average MAPE and CV-RMSE are 0.8564, 1.975\% and 0.0687\%, respectively. The MAPE of SVR model is the largest among all of the models; an average error is about 2.3341\%. On the other hand, the CV-RMSE of SVR model is the largest among all models; an average error is about 0.0809\%.  According to the MSE, average MAPE and CV-RMSE values, the electricity consumption forecasting accuracy of tested models in descending order is as follows: RCNN-SVR, RCNN, MPSO-BP, DeepEnergy, and SVR. However, SVR uses less time than other models (1.82 s), comparing with the rest three methods, our model RCNN-SVR uses relatively less time than RCNN, SVR and DeepEnergy methods. 

\begin{figure}[H]
\centering
\includegraphics[scale=0.45]{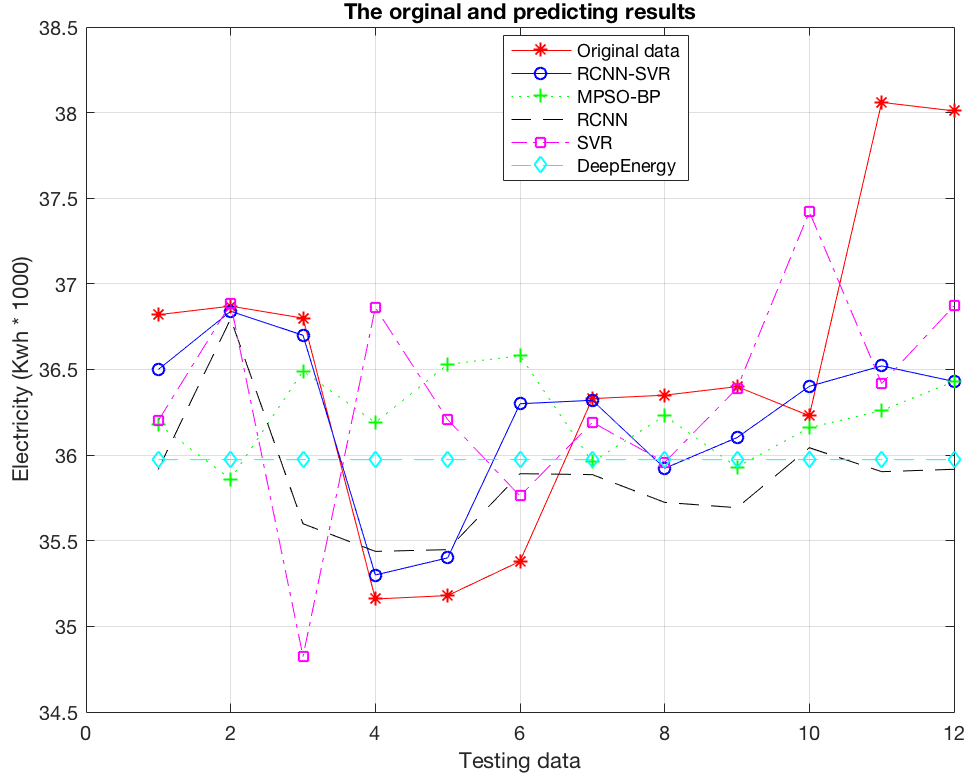}\\
\caption{The comparison results of predicting electricity values using RCNN-SVR, RCNN, SVR, MPSO-BP and DeepEnergy.  } 
\end{figure}

\begin{figure}[H]
\centering
\includegraphics[scale=0.5]{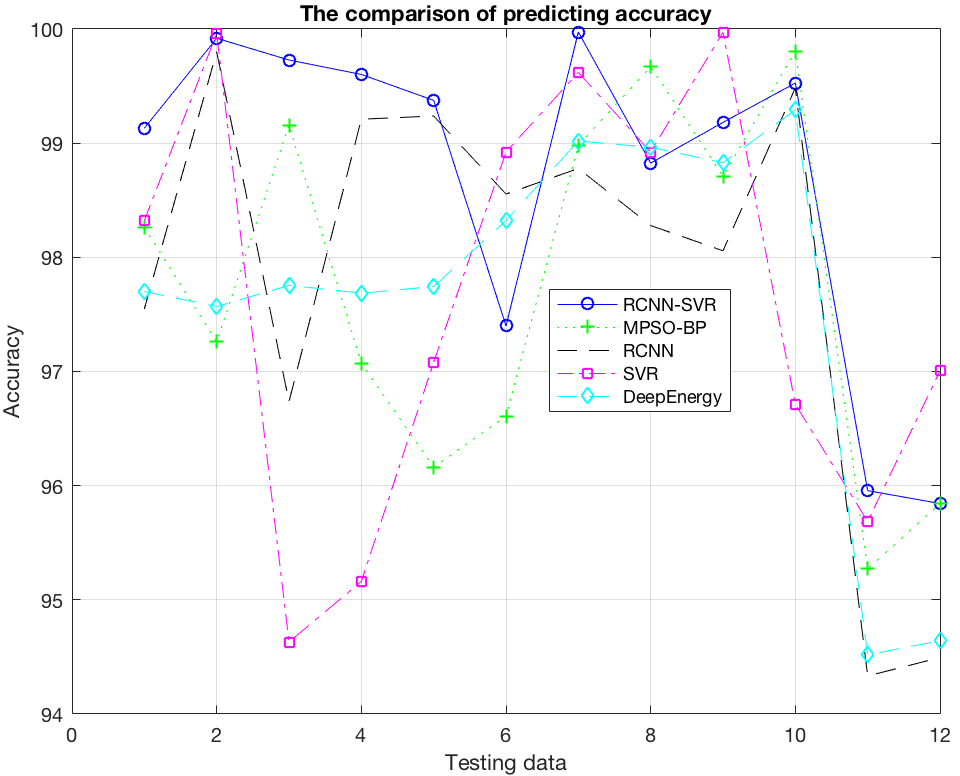}\\
\caption{The comparison results of predicting accuracy using RCNN-SVR, RCNN, SVR, MPSO-BP and DeepEnergy. SVR model is more likely to oscillate in the image which means the predicting results are not stable. And RCNN-SVR is more closed to the true data. Although DeepEnergy network have better accuracy in Fig. 8, but its actually predicted values not change with different testing data and this is caused by too many layers in the DeepEnergy network with limited training data (in Fig.7). }  
\end{figure}

\begin{table}[hbt!]
\setlength{\tabcolsep}{3.5pt}
\caption{Predicting results of the electricity consumption by different methods } \label{tab:my_label}
\centering
\begin{tabular}{cccccc}
\hline
	& \textbf{RCNN-SVR}	& \textbf{RCNN} & \textbf{SVR} & \textbf{MPSO-BP \cite{Zhang}} &\textbf{DeepEnergy \cite{Ping}}\\
\hline
MSE		& \bf{0.8564}	& 1.0690 & 1.1639 & 0.9236 & 1.0720\\
MAPE 	& \bf{1.975\%}	& 2.1239\% & 2.3341\% & 2.2665\% & 2.330\% \\
CV-RMSE & \bf{0.0687\%}  & 0.0755\% & 0.0809\%   &0.0745\% &0.0760\% \\
Time (s)	& 4.35		& 221.74 & \bf{1.82} & 27.21 & 3758\\
\hline
\end{tabular}
\end{table}

From table \ref{tab:my_label}, we can find that our RCNN-SVR model has the smallest MSE, MAPE, and CV-RMSE, which means our model has the highest accuracy than other methods. Therefore, the RCNN-SVR model is the most suitable method for electricity predicting. We recommend using the RCNN-SVR model to predict the electricity consumption of mineral company.

\subsection{Forecasting of electricity consumption of each month in 2018}

Using the trained RCNN-SVR model, we predict the electricity consumption values of each month in 2018, as shown in table \ref{tab:my_label2}. The electricity consumption will increase in November and December, and this may cause by heavy pressure on the operation, maintenance, and supply heating of power system.

\begin{table}[!hbtp]
\setlength{\tabcolsep}{8.5pt}
\caption{Forecasting electricity consumption (kWh) of each month in 2018 }\label{tab:my_label2}
\centering
\begin{tabular}{ccccccc}
\hline
\textbf{Months}	& \textbf{1}	& \textbf{2} & \textbf{3} & \textbf{4}& \textbf{5}& \textbf{6}\\
\hline
Evs (kwh/t)    &38.78    &38.56    &37.01   & 36.83    &35.76    &35.94 \\
\hline
\textbf{Months}	& \textbf{7}& \textbf{8}& \textbf{9}& \textbf{10}& \textbf{11}& \textbf{12}\\
\hline
Evs (kwh/t)     &37.33    &36.84    &36.76    &36.06    &37.78    &38.28 \\
\hline
\end{tabular}
\end{table}

\section{Discussion}

The traditional method, such as SVR, BP neural network has been applied in electricity consumption prediction. In this paper, these methods also provided a reasonable result (as shown in table 1). Regarding SVR, the results are worst among these methods. One reason is that there are no enough features can be trained due to the limited data. According to the table \ref{tab:my_label}, the RCNN has a relative long computational time, and this is caused by the features extraction and training step. Our RCNN-SVR model has the lowest MSE, MAPE, and CV-RMSE comparing with other methods. Furthermore, the selection of extracting features from which layer in RCNN-SVR model is important, as shown in Fig. 9, MSE is overall reduced with the selected later layers. And this implies that the most useful features are shown in the last layers in the RCNN-SVR model. Therefore, we may get better results if we use the features from the last layer.

\begin{figure}[H]
\centering
\includegraphics[scale=0.5]{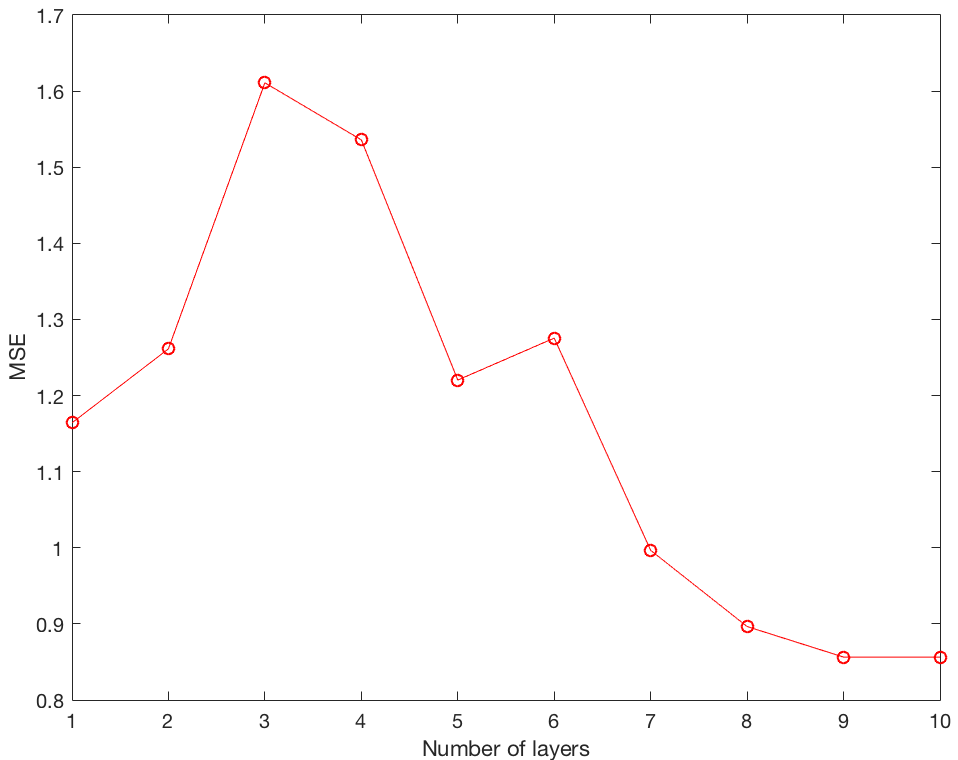}\\
\caption{The relationship between MSE and number of layer selected in RCNN-SVR model. }
\end{figure}

\section{Conclusions} \label{sec:conclusion}
In this paper, we propose a regressive convolution neural network and support vector regression (RCNN-SVR) model for electricity consumption forecasting. The proposed model is validated by experiment with the electricity consumption data from the past
five years. In the experiment, the data from a mineral company were used, and historical electricity demands are considered. According to the experimental results, the RCNN-SVR model can precisely predict electricity consumption in the next following months. Also, the proposed model is compared with four models that were used in electricity consumption forecasting. The comparison results showed that performance of our RCNN-SVR model is the best among all tested algorithms, which has the lowest values of MSE, MAPE, and CV-RMSE. According to all of the obtained results, the proposed method can reduce computation time. The proposed RCNN-SVR method successfully solves three issues which are mentioned above: (1) reduce the computational cost; (2) train the model with  limited  data;  and  (3)  improve  the  prediction  accuracy. Therefore, the RCNN-SVR  model  can be used to  predict  the  electricity  consumption  of  mineral company.

However, our paper has the limitation of data size. For future work, we will first test our model use more data, then we will expand the different neural networks, such as DenseNet, Adversarial neural network to extract the features of data. What's more, the novel model in this paper can be used in predicting electricity values in other fields, such as wind power generation system electricity prediction, and agricultural electricity consumption area.

\end{document}